\documentclass[review]{elsarticle}

\journal{}

\usepackage{lineno,hyperref}
\modulolinenumbers[5]

\usepackage{url}
\usepackage{latexsym}
\usepackage{graphicx}
\usepackage[tight,footnotesize]{subfigure}
\usepackage{placeins}
\usepackage{stfloats}
\usepackage{pgfplots}
\usepackage[export]{adjustbox}
\usepackage{array}
\usepackage{verbatim}
\usepackage{flushend}
\usepackage{picins}

\newcommand{\mysubsection}[1]{\vspace{0.3em} \noindent\textbf{#1}}

\makeatletter
\def\ps@pprintTitle{%
   \let\@oddhead\@empty
   \let\@evenhead\@empty
   \let\@oddfoot\@empty
   \let\@evenfoot\@oddfoot
}
\makeatother

\begin{document}
\begin{frontmatter}

\title{A Comparative Analysis of Content-based\\ Geolocation in Blogs and Tweets}

\author{Konstantinos~Pappas\corref{cor1}}
\ead{pappus@umich.edu}

\author{Mahmoud~Azab\corref{cor2}}
\ead{mazab@umich.edu}

\author{Rada~Mihalcea\corref{cor3}}
\ead{mihalcea@umich.edu}

\address{University of Michigan}
\address{2260 Hayward street, Ann Arbor, MI, 48109, USA}

\begin{abstract}
The geolocation of online information is an essential component in any geospatial application. While most of the previous work on geolocation has focused on Twitter, in this paper we quantify and compare the performance of text-based geolocation methods on social media data drawn from both Blogger and Twitter. 
We introduce a novel set of location specific features that are both highly informative and easily interpretable, and show that we can achieve error rate reductions of up to 12.5\% with respect to the best previously proposed geolocation features. We also show that despite posting longer text, Blogger users are significantly harder to geolocate than Twitter users. Additionally, we investigate the effect of training and testing on different media (cross-media predictions), or combining multiple social media sources (multi-media predictions). Finally, we explore the geolocability of social media in relation to three user dimensions: state, gender,  and industry.
\end{abstract}


\end{frontmatter}

\section{Introduction}
\label{sec:intro}

There is an ever-growing amount of online information, which among other things has also led to a proliferation of geospatial technologies. The construction of models  that can accurately predict users' locations has been identified as a priority in supporting geospatial applications such as the improvement of local search tools  \cite{bouidghaghen11}, event detection \cite{Weng11,Li12}, disaster response management \cite{Latonero11,Earle12}, targeted advertising \cite{Wanek11}, defense and security applications \cite{Yang12,Kandias13}.

Despite this need, and despite the increased availability of embedded GPS technologies and geotagging capabilities offered by many online applications,  only a small number of users disclose their actual location ~\cite{Cheng10,Abrol2012,Stefanidis13,Dredze2013}. To complete the missing location information and to better support geospatial applications, researchers have proposed a number of location detection methods based on the content generated by the users \cite{Eisenstein10,Mahmud12,Han14,Liu15}. 

Most of this previous content-based geolocation work has however exclusively focused on one social media source, namely Twitter. In this paper, we address this shortcoming of previous research and perform extensive evaluations and comparisons using two social media streams: blogs and tweets.

While Twitter has clearly dominated the past ten years of text-based geolocation research, we show that this prior work on Twitter does not perform analogously on Blogger, both quantitatively (prediction accuracy and methods) and qualitatively (user locatability). We introduce alternative geolocation methods that are substantially more efficient than previous work, in terms of performance, execution time, and interpretability.

The paper makes four main contributions. First, we build two large comparable datasets of blogs and tweets, consisting of users with a known U.S. state location, which allows us to draw comparisons between text geolocation in these two social media. 

Second, we advance the state-of-the-art in manual feature engineering for geolocation prediction. We propose two new feature selection strategies that lead to classification results significantly exceeding the results obtained with feature selection methods from previous work. As an additional advantage, the features that are selected with our  methods are not only location specific and  concise, but also easy to interpret.

Third, we perform comparative evaluations of geolocation classifiers at state level on both blogs and tweets, and highlight the differences between these two media. Moreover, we further explore these differences in cross-media and multi-media geolocation predictions, and show the effect of training on different media or a mixed media dataset. To our knowledge, this is the first time that such a comparison for geolocation prediction in different social media has ever been made.

Finally, we analyse the relation between geolocatability and three demographic dimensions:  state,  gender, and  industry. We show that these user properties are related to the accuracy of geolocation classifiers, with certain states/genders/industries being easier to geolocate than others.

Note that in our work we only make use of the text generated by social media  users (i.e., blog posts or tweets), and do not rely on additional profile information (except for the geolocatability analyses). While previous research has found that location-related profile fields (i.e., city, country, state, and country) can help the geolocation prediction \cite{Dredze2013,Han14}, we focus our analysis on geolocating users based on their posted text alone, targeting the more challenging and frequent scenario where explicit location-related profile information is absent.

After discussing related work in Section \ref{sec:related}, we present the datasets used in this study in Section \ref{sec:datasets}. Section \ref{sec:geolocation} presents the evaluation of different features selection methods and introduces a novel set of features, the lexicons. We measure the performance of cross-media predictions and augmenting the training set using data from both platforms in Section \ref{sec:crossgeo}. We evaluate how the different profile metadata correlate with users' locatability and expose which population subgroups are easier to geolocate in Section \ref{sec:meta}, and finally discuss our findings and conclude in Section \ref{sec:conclusion}.

\section{Related Work}  
\label{sec:related}

Previous work on geolocation can be grouped into three broad categories. The first type relies on network infrastructure, and use geolocation databases to map the IP address of the users  to their geographic location \cite{Eriksson10,Poese2011}. Another set of approaches make use of social network relations and geolocate social media users based on their \textit{friend} or \textit{follow} relations
\cite{Backstrom10,Davis11,Sadilek12,Rout2013,Jurgens15}; the intuition here is that frequent interactions tend to occur between users with close geographic proximity. Finally, the third category of methods, also endorsed in this paper, relies on the textual content generated by social media users. In this section, we  review this latter type of approaches.

While one of the earliest content-based geolocation studies sought to determine the geographical focus based on the toponyms mentioned in blogs \cite{Fink09}, most of the subsequent work focused on Twitter datasets \cite{Cheng10,Eisenstein10}.

Following those initial efforts, Wing and Baldridge \cite{Wing11} attempt geodesic grids classifications using supervised models and Hecht et al. \cite{Hecht11} define the \textit{CALGARI} algorithm to predict the users' country and state. Similarly, Kinsella et al. \cite{Kinsella11} classify at country and zip code granularity using the Ponte and Croft \cite{Ponte98} approach to build models of location, while Chang et al. \cite{Chang12} try unsupervised models and 100-miles radius regions. Other classifiers have also been tried, including K-Nearest Neighbor \cite{Roller12} and ensembles of classifiers \cite{Mahmud12}.

In the following years, rather than changing the classifiers, research has focused on generative models using location indicative words identified via feature selection \cite{Han12}, or using  user metadata \cite{Han13}. They evaluate their approach on a number of metrics including accuracy, 100-miles radius ``near-miss" accuracy, mean, and median prediction error. 

On other types of social media, Popescu and Grefenstette \cite{Popescu10} analyze the tags on Flickr photos to infer the users' location and gender, and Wing and Baldridge \cite{Wing14} use data from Twitter, Wikipedia, and Flickr to create a model based on logistic regression and geotag text to grid granularity similarly to Roller et al. \cite{Roller12}. Finally, Rahimi et al. \cite{Rahimi15} combine the network- and text- based methods into a hybrid approach that uses logistic regression and label propagation; they measure the 100-mile accuracy, mean, and median error on three different Twitter datasets. Similar hybrid approaches leverage Graph Convolutional Networks \cite{Rahimi18} and Gaussian mixture models \cite{Bakerman18} to further increase geolocation performance.

Previous work has verified that simple generative models with appropriate feature engineering can indeed outperform more sophisticated methods \cite{Priedhorsky14,Han14}, including deep learning \cite{Liu15} \footnote{Some of the most recent deep learning attempts yield promising performance \cite{Rahimi17,Lourentzou17}, but these results are inherently more challenging to analyze and interpret, more expensive to acquire (computationally, time-wise as well as optimizing the architecture) and neural nets are extremely data hungry (customarily requiring millions of examples) making them less attractive in qualitative studies especially when targeting text from social media that are less prevalent than Twitter where data is not so abundant.}.
In this paper, we adopt this guideline and introduce new feature weighting and selection methods that improve both the accuracy and effectiveness of geolocation algorithms. Furthermore, unlike most previous research, we target both a blogging and a microblogging platform, and examine individual, mixed and cross-media geolocation performance.

\section{Datasets}
\label{sec:datasets}

We use two corpora collected from two widely used social media platforms, Blogger and Twitter, geolocated at state-level.
Our decision to focus on state-level geolocation is motivated by recent previous work that used a similar location granularity \cite{Eisenstein10,Liu15}, as well as by the lack of availability of geo-coordinates in blog data (only about 0.5\% of the blog posts include such geospatial information). Note however that our methodology is not restricted to state-level geolocation, and it could be generalized to the prediction of finer-grained locations such as cities \cite{Han12} or hierarchically structured grid cells \cite{Wing14}.

To control for the distribution differences of users in the two social media platforms both datasets include the same number of users (56,750),  equally distributed across the 50 U.S. states. In our experiments, we randomly split each dataset in a train, a development and a test set, with 45,350, 5,700 and 5,700 users respectively.

\subsection{U.S. Blogs}
\label{subsec:blogger}

Our goal is to build a large dataset of geolocated blogs with U.S. state information. We first start by collecting a set of profiles for bloggers that meet our location specifications, by  searching for individual states on the profile finder on \url{http://www.blogger.com}. Note that the profile finder only identifies users that have an exact match with the location specified in the query; we thus run queries that use both the state abbreviations (e.g., TX, AL), as well as the state full names (e.g., Texas, Alabama). We then apply three data filtering steps: we exclude all the group blogs, which do not have individual profile elements; we also exclude all the blogs that have no associated blog posts;  and we exclude all the profiles whose cumulative posted textual content is less than 600 characters.

After all the processing steps, we collect 56,750 Blogger users with state location information equally distributed across the U.S. states (1,135 users per state). For each of these bloggers, we find their blogs (a blogger can have multiple blogs), for a total of 95,217 blogs. For each of these blogs we identify the 21 most recent blog posts,\footnote{Both our datasets were collected in summer/fall 2015.} which are cleaned of HTML tags, finally resulting in a collection of 1,283,521 blog posts. Unlike tweets, which represent the other popular social media stream, we find that blog posts are significantly longer than 140 characters (the maximum length of a tweet).Table \ref{table:statisticsBlogger} shows the maximum, mean, standard deviation, and median number of blogs and characters.

The final processing step is the tokenization of the blog posts, performed using the Stanford tokenizer \cite{Manning14}.

\begin{table}[htbp]
\centering
\scalebox{0.8}{
\begin{tabular}{ l r r r r r }
\hline
& Max & Mean & $\sigma$ & Median \\
\hline
blogs\\
per user & 99 & 1.68 & 2.21 & 1\\
\hline
blog posts\\
per user & 1075 & 22.62 & 24.58 & 21\\
\hline
characters\\
per post & 889,587 & 2,044 & 4,152 & 1,104\\
\hline
characters\\
per user & 15,265,769 & 46,274 & 110,253 & 27,026\\
\hline
\end{tabular}
}
\vspace{0.15in}
\caption{Statistics on the Blogger dataset.\label{table:statisticsBlogger}}
\end{table}

\mysubsection{Blog Metadata.} Blogger profiles are accompanied by a rich set of metadata, including fields such as city, occupation, industry, interests, movies, etc. which can be very useful in studies that connect words with demographics \cite{Garimella17,Pappas16A,Pappas16B}.
However, except for the city field, which naturally leads to a high geolocation accuracy (58.8\%) 
when incorporated as a feature, the information provided by all the other fields gives consistently low performance in the geolocation task using the classifiers described in the following section, with accuracy figures below 9.3\%. Throughout this paper, as mentioned in the introduction, we therefore focus on geolocation based on the content of the blog posts, and ignore the metadata. 

\subsection{U.S. Tweets}
\label{subsec:twitter}

In addition to the Blogger dataset, we also collect a Twitter dataset that emulates the statistics of the blog dataset, making them directly comparable. Similar to Blogger, we only consider Twitter users whose location profile field matches either a state's abbreviation or a state's full name.

Starting with a user's ID, we download their most recent 200 tweets. We remove all the retweets, and as done in previous work  we exclude all the mentions and hashtags \cite{Han14}. After all these processing steps, we only keep the users that have a total of at least 600 characters. To match the distribution of the Blogger dataset, we collect 1,135 users per state, for a total of 56,750 users.  Table \ref{table:statisticsTwitter} shows the  statistics of our Twitter dataset\footnote{Although the maximum length of a tweet is restricted to 140 characters, in our dataset we find the maximum tweet length to be 509 characters. This happens because the Twitter API uses the HTML representation for all the symbols (e.g., the symbol `$>$'  is represented by four characters: `\&gt;').}. 

As a last processing step, all the tweets are tokenized using a version of a regex tokenizer specifically designed for Twitter \cite{Han14}.

\begin{table}[htbp]
\centering
\scalebox{0.8}{
\begin{tabular}{ l r r r r r }
\hline
& Max & Mean & $\sigma$ & Median \\\hline
tweets\\
per user  & 659 &  143.7 & 52.48 & 156 \\
\hline
characters\\
per tweet  & 509 &  83.49 & 37.7 & 85 \\
\hline
characters\\
per user  & 88,351 & 12,487.98 & 6,266.11 & 12,316.5\\
\hline
\end{tabular}
}
\vspace{0.1in}
\caption{Statistics on the Twitter dataset.\label{table:statisticsTwitter}}
\end{table}

\mysubsection{Twitter Metadata.}
Previous geolocation studies have found various Twitter profile metadata such as the timezone and the declared location of the user to be informative \cite{Han14}. However, since we want to explicitly focus on content-based geolocation, throughout the experiments reported in this paper we ignore the profile metadata.

\section{Content-based Geolocation}\label{sec:geolocation}

We approach the geolocation task in three main steps. First, we filter the input text, and remove words unlikely to help the classification task. Second, we weight the features, and select a subset of the features based on their weight or based on other heuristics. Finally, we use a machine learning classifier to predict the most likely state.

\subsection{Pre-filtering}

Previous studies on geolocation disregarded words that included non-alphabetic characters, were less than three characters long, or had a frequency less than 10, as they were considered to be  ``low-utility" words \cite{Han14}. However, other studies have found that short words (e.g., LA, NY, DC), street names (e.g., 74th), area codes, and street numbers 
can have geolocation value ~\cite{Chang12,Roller12}. Therefore, we only exclude words that are rare among the users in the training set, assuming they will be infrequently used by other users as well. Examples of such words are URLs, typos, rare names, and different variants of punctuation symbols.
Following this intuition, we define our filter to only consider words that appear in the text of at least three different users from the training corpus. This leaves us with a total of 317,027 different words in our Blogger training set, and 150,244 different words in our Twitter training set.

\begin{figure*}[tp]
\small
\begin{tabular*}{1\textwidth}{@{\extracolsep{\fill}} c c c }
\multicolumn{1}{c}{\includegraphics[width=0.30\textwidth]{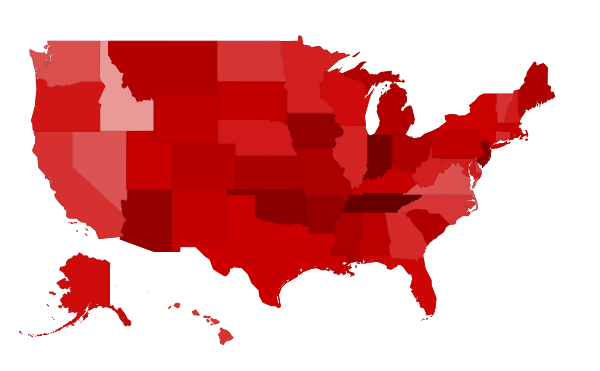}} &\multicolumn{1}{c}{\includegraphics[width=0.30\textwidth]{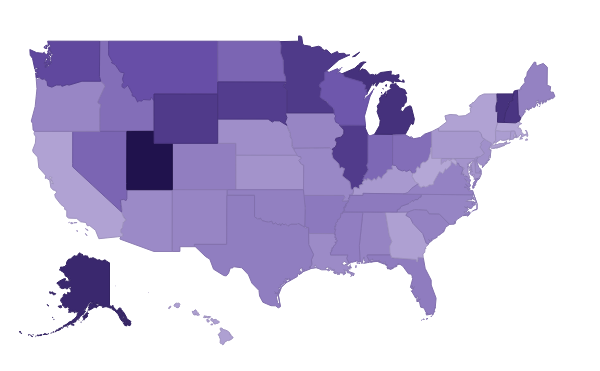}} &\multicolumn{1}{c}{\includegraphics[width=0.30\textwidth]{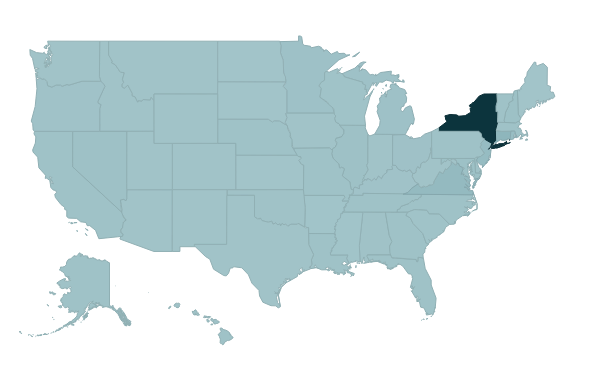}}\\
\multicolumn{1}{c}{today} & \multicolumn{1}{c}{lake} & \multicolumn{1}{c}{bronx}\\
\end{tabular*}
\caption{Distribution of three selected words across the 50 U.S. states.} 
\label{fig:bWordsDist}
\end{figure*}

\subsection{Feature Weighting and Selection}
\label{subsec:lex}

Our premise is that we can exploit certain aspects of the geographical variability of language to construct improved geolocation models. This intuition is based on the observation that the proportional frequency of certain words changes for different U.S. states. 
To illustrate, consider the geographical distribution for three words, as shown in Figure \ref{fig:bWordsDist}.\footnote{In all the maps we generate, the darker the color of a state, the higher the proportion of instances in that state that match the criterion used to generate the map.} For each state, we measure the frequency of the selected word and divide it by the sum of the frequencies of all the  words in that state. 
For example, ``today" constitutes an instance of a \textit{common} word: its relative appearance is nearly constant across all states. In contrast, ``lake" is discernibly used more in northern states. We believe that even though such words, when present, can be valuable in predicting a user location, they are still orders of magnitude less revealing than 1-local words. To emphasize this point, we also map the relative appearance of the word ``bronx" which clearly indicates the NY state.

In line with this intuition, we implement and test several feature selection approaches, which aim to narrow down the vocabulary to those words that are most useful for the task of geolocation. Aside from increased accuracy, as shown in the results below, such feature selection strategies also have the effect of increasing the efficiency of the classification algorithm, as it now has to deal with a significantly smaller number of features.

\mysubsection{Information gain ratio (IGR).} The IGR represents the state-of-the-art in terms of manual feature selection methods for the purpose of geolocation  ~\cite{Han14}. The IGR of a word {\it w}, across all states {\it S}, is defined as the ratio between its information gain value {\it IG}, which measures the decrease in class entropy {\it H} that {\it w} brings, and its intrinsic entropy {\it IV}, which measures the entropy of the presence versus the absence of that word:

\vspace{1mm}

\[\resizebox{0.7\textwidth}{!}{$IGR(w) = \frac{IG(w)}{IV(w)} \propto \frac{-H(S|w)}{-P(w)logP(w)-P(\overline{w})logP(\overline{w})} \propto $}\]

\[\resizebox{0.7\textwidth}{!}{$ \frac{P(w)\sum_{s \in S}P(s|w)logP(s|w)+P(\overline{w})\sum_{s \in S}P(s|\overline{w})logP(s|\overline{w})}{-P(w)logP(w)-P(\overline{w})logP(\overline{w})} $}\]

\vspace{4mm}

A weakness of this measure is the fact that it ranks each word depending on whether its appearance reduces the entropy across all the states, which does not align well with our goal of identifying words that unambiguously hint to only one location. Despite this drawback, to facilitate a comparison with earlier work, we also implement and test the IGR feature selection method.

\mysubsection{Word locality heuristic (WLH).} WLH is a heuristic that we introduce, which promotes words primarily associated with one location (i.e., one U.S. state, in our case). We first measure the probability of a word occurring in a state, divided by its probability to appear in any state. Then, for a given word \textit{w}, we define the \textit{WLH} as the maximum such probability across all the states \textit{S}:
\[WLH(w) = \max_{s \in S}{\frac{P(w|s)}{P(w)}}\]

\mysubsection{Location lexicons.} Identifying words that are strongly associated with one location is an effective ranking scheme. However, this alone does not alleviate the massive number of features that inhibits the use of discriminative classifiers. To address this issue, we also extend our WLH method by grouping the location-specific words into class-dependent sets. Specifically, for each prediction class (i.e., U.S. state) we create a lexicon that contains the most significant words for that class. We adhere to three rules when building these lexicons: (1) we keep only words that are used by at least a \textit{p} number of users; (2) we include only words that have a WLH score above a certain threshold \textit{h}; and (3) we enforce that each lexicon contains at least \textit{t} words.

The intuition behind these parameters is as follows. The first parameter, \textit{p}, ensures that the words included in the lexicons are used by many users and hence, have a high chance of appearing in the text of future users. The second parameter, \textit{h}, ensures that only words that are highly indicative of a location are included. The third parameter, \textit{t}, denotes the smallest allowed size of a lexicon. This last restriction ensures that no lexicon is left empty, in which situation some states would not have any representative word making it impossible to classify any future text to them. If the \textit{t} threshold is not met for a lexicon, we relax the \textit{h} score restriction in order to allow more words to be included in that lexicon.

\begin{table}[bh!]
\centering
\scalebox{0.8}{
\begin{tabular}{  l | c }
\hline
\textbf{State (media)} & \textbf{Lexicon}\\
\hline
CA (Blogger) & kat, pe, commerce\\
MI (Blogger) & arbor, amp, michigan\\
NY (Blogger) & headlines, prediction, provision\\
TX (Blogger) & tx, austin, houston\\
\hline
CA (Twitter) & ca, francisco, oakland \\
MI (Twitter) & detroit, michigan, mi\\
NY (Twitter) & ny, brooklyn, york\\
TX (Twitter) & tx, houston, austin, dallas\\
\hline
\end{tabular}
}
\vspace{0.1in}
\caption{Sample words in the state lexicons.\label{table:lexicons}}
\end{table}

Sample words from the state lexicons derived from blogs or tweets are presented in Table \ref{table:lexicons}. While location names are generally common in these  lexicons, the blog lexicons also have exceptions, e.g., popular states such as NY, which are highly-populated and diverse and for which location words tend to be less informative.

\begin{figure}
\begin{center}

\scalebox{0.8}{
\begin{tikzpicture}
\begin{axis}[
xlabel=Lexicons' Size, ylabel=Jaccard Coefficient
]
\addplot[color=red,mark=x]
coordinates { (100,0.149425287356321853) (200,0.19161676646706588) (300,0.1650294695481336) (400,0.14244186046511628) (500,0.128)
(600,0.11393596986817325) (700,0.10581583198707593) (800,0.0992957746478873) (900,0.09606986899563319) (1000,0.08924611973392461)};
\end{axis}
\end{tikzpicture}
}
\caption{Blogger and Twitter lexicons' overlap.\label{fig:overlap}}
\end{center}
\end{figure}
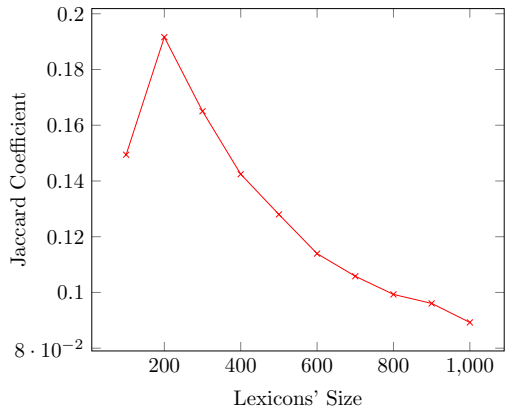

Interestingly, the lexicons generated for the two social media have only little overlap, as shown in Figure \ref{fig:overlap}, which plots the Jaccard coefficient for Blogger and Twitter lexicons as a function of the lexicons size. This suggests that there are significant differences between the location indicative words used in the two media.

\subsection{Geolocation Classifiers}
\label{sec:detection}

Using our two datasets of blogs and tweets, and using the feature selection methods described in the previous section, we run several comparative experiments. We use a multinomial Naive Bayes (NB) classifier, as done in previous work \cite{Han14,Wing14}, as well as an SVM classifier \cite{Vapnik95, Fan08}.\footnote{We use the NB classifier as implemented in Weka, the LibSVM classifier with a linear kernel, and LibLinear.} 

As a baseline, we implement an NB classifier that uses all the words as features. We find that this baseline yields significantly different results in the two media, 9.3\% for blogs, and 28.53\% for tweets, which suggests a difference in the user geolocatability of these two sources.

We also experiment with a word embedding representation, where we use word vectors obtained with GloVe \cite{Pennington14} trained on a Common Crawl and a Twitter dataset respectively, which are added up and averaged to create a word vector representation for each user in our data. However, preliminary experiments using this approach did not show promise, with accuracy figures of 7.2\% for Blogger and 17.3\% for Twitter in our test sets.  

Recall from Section \ref{sec:datasets} that we work with two corpora, one consisting of blogs and one consisting of tweets, both including 56,750 users equally distributed across the 50 U.S. states. Each dataset is split into a training set of 45,350 users, a development set of 5,700 users, and a test set of 5,700 users.

\begin{figure*}[htbp]
\begin{center}
\begin{tikzpicture}

\begin{axis}[
  small,
  title=Blogger,
  title style={font=\fontsize{10}{12.5}\selectfont},
  name=blogger,
  legend columns=2,
  legend pos=south east,
  legend style={at={(0.5,0.94)}, anchor=north, font=\fontsize{7}{8.5}\selectfont},
  legend image post style={scale=1},
  height=0.6*\textwidth,
  width=1*\textwidth,
  enlargelimits=0.02,
  xtick={0,10,20,30,40,50,60,70,80,90,100},
  ytick={0,10,20,30,40,50,60},
  ylabel={Accuracy (\%)},
  xlabel={Attributes (\%)},
  ymin=0,
  ymax=60
  ]
  \addplot[mark=star, red] coordinates { (0.5,6.59) (1,9.9) (1.5,11.83) (2,13.53) (2.5,14.65) (3,15.51) (4,17.79) (5,19.35) (10,24.56) (15,27.58) (20,28.86) (25,28.97) (30,30.47) (35,32.7) (40,32.97) (45,33.88) (50,34.19) (55,34) (60,34.58) (65,34.16) (70,33.93) (75,34.02) (80,33.16) (85,32.21) (90,29.86) (95,28.07) (100,22.88) };
  \addlegendentry{WLH}

  \addplot[mark=diamond, blue] coordinates { (0.5,8.7) (1,10.53) (1.5,10.97) (2,12.37) (2.5,12.89) (3,13.23) (4,13.84) (5,13.97) (10,15.83) (15,17.23) (20,17.95) (25,18.56) (30,19.09) (35,19.4) (40,19.98) (45,20.05) (50,20.46) (55,21.33) (60,21.63) (65,22.91) (70,23.61) (75,25.12) (80,28.07) (85,30.79) (90,30.35) (95,28.39) (100,23.28) };
  \addlegendentry{IGR(best)}

  \addplot[mark=triangle, orange] coordinates { (0.5,8.65) (1,10.53) (1.5,10.97) (2,12.37) (2.5,12.89) (3,13.23) (4,13.84) (5,13.97) (10,15.56) (15,17.02) (20,17.56) (25,18.23) (30,18.28) (35,18.58) (40,19.4) (45,19.09)  (50,19.3) (55,19.7) (60,19.86) (65,20.33) (70,20.58) (75,21.28) (80,22.9) (85,23.83) (90,15.75) (95,13.39) (100,9.25) };
  \addlegendentry{IGR(NB)}

  \addplot[mark=none, brown, very thick] coordinates { (0,32.6) (100,32.6) };
  \addlegendentry{Lexicons}
  
  \addplot[mark=none, yellow, very thick] coordinates { (0,22.44) (100,22.44) };
  \addlegendentry{All Words}
\end{axis}

\end{tikzpicture}
\end{center}
\caption{Geolocation accuracy for the different feature selection methods in Blogger.} 
\label{fig:featSelB}
\end{figure*}
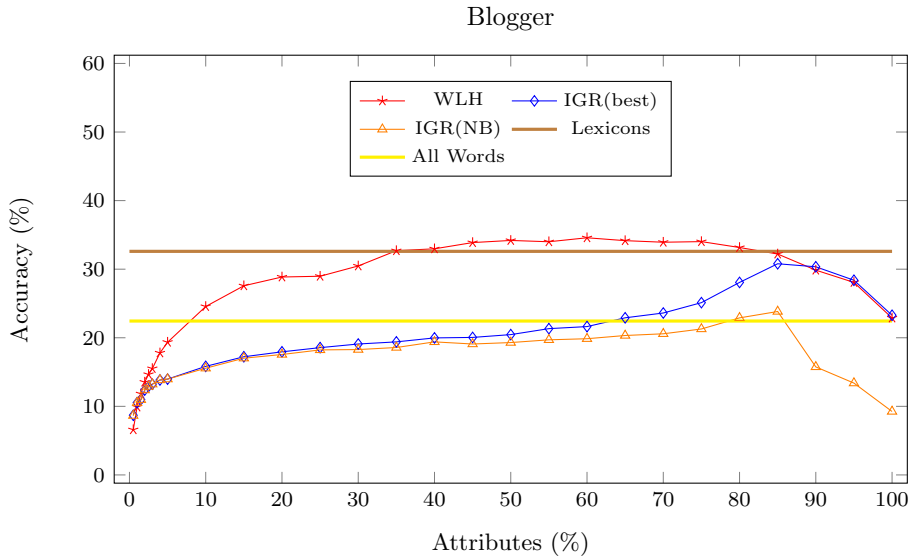

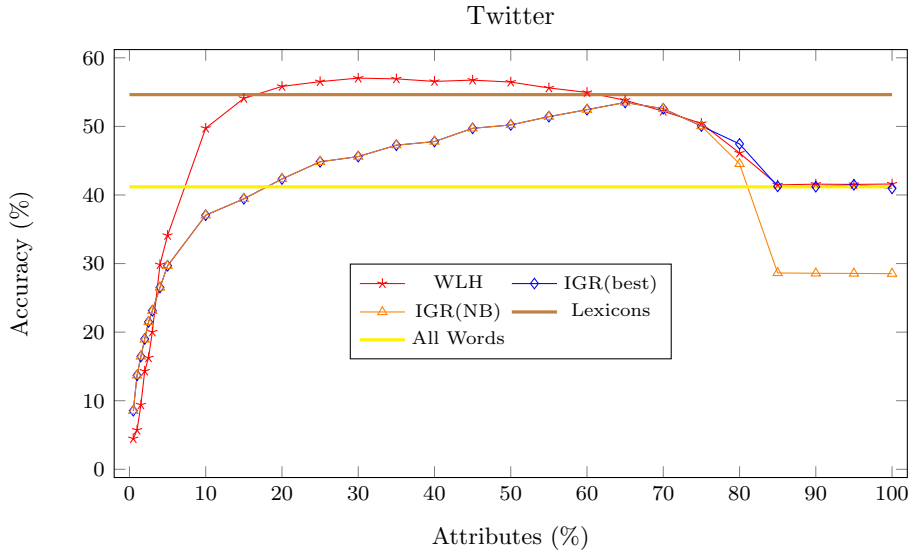
\begin{figure*}[htbp]
\begin{center}
\begin{tikzpicture}

\begin{axis}[
  small,
  title=Twitter,
  title style={font=\fontsize{10}{12.5}\selectfont},
  name=twitter,
  xshift=10mm,
  at=(blogger.right of south east),
  anchor=left of south west,
  legend columns=2,
  legend pos=south east,
  legend style={at={(0.5,0.5)}, anchor=north, font=\fontsize{7}{8.5}\selectfont},
  legend image post style={scale=1},
  height=0.6*\textwidth,
  width=1*\textwidth,
  enlargelimits=0.02,
  xtick={0,10,20,30,40,50,60,70,80,90,100},
  ytick={0,10,20,30,40,50,60},
  ylabel={Accuracy (\%)},
  xlabel={Attributes (\%)},
  ymin=0,
  ymax=60
  ]
  \addplot[mark=star, red] coordinates { (0.5,4.47) (1,5.72) (1.5,9.4) (2,14.32) (2.5,16.26) (3,20.02) (4,29.84) (5,34.11) (10,49.74) (15,54.09) (20,55.83) (25,56.53) (30,57.04) (35,56.93) (40,56.56) (45,56.75) (50,56.46) (55,55.61) (60,54.95) (65,53.81) (70,52.21) (75,50.44) (80,46.12)
  (85,41.47) (90,41.58) (95,41.54) (100,41.6) };
  \addlegendentry{WLH}
  
  \addplot[mark=diamond, blue] coordinates { (0.5,8.54) (1,13.7) (1.5,16.46) (2,19) (2.5,21.49) (3,23.12) (4,26.51) (5,29.68) (10,37.05) (15,39.44) (20,42.33) (25,44.84) (30,45.6) (35,47.26) (40,47.79) (45,49.72) (50,50.21) (55,51.42) (60,52.44) (65,53.49) (70,52.58) (75,50.05) (80,47.44)
  (85,41.26) (90,41.21) (95,41.49) (100,40.97) };
  \addlegendentry{IGR(best)}

  \addplot[mark=triangle, orange] coordinates { (0.5,8.54) (1,13.7) (1.5,16.46) (2,18.93) (2.5,21.49) (3,23.12) (4,26.51) (5,29.68) (10,37.05) (15,39.44) (20,42.33) (25,44.84) (30,45.6) (35,47.26) (40,47.79) (45,49.72) (50,50.21) (55,51.42) (60,52.44) (65,53.49) (70,52.58) (75,50.05) (80,44.54) (85,28.63) (90,28.58) (95,28.56) (100,28.53) };
  \addlegendentry{IGR(NB)}
  
  \addplot[mark=none, brown, very thick] coordinates { (0,54.63) (100,54.63) };
  \addlegendentry{Lexicons}
  
  \addplot[mark=none, yellow, very thick] coordinates { (0,41.19) (100,41.19) };
  \addlegendentry{All Words}
  
\end{axis}
\end{tikzpicture}
\end{center}
\caption{Geolocation accuracy for the different feature selection methods in Twitter.} 
\label{fig:featSelT}
\end{figure*}

\mysubsection{Experiments on development data.} Figures \ref{fig:featSelB} and \ref{fig:featSelT} show the performance of the different feature selection methods as obtained on the blog and tweet development datasets.  For IGR and WLH, we plot the accuracy achieved for different percentages of features used. For ``Lexicons", we  use all the features available, and therefore represent the accuracy as a straight line to allow for easy comparison with the other two methods.
We also implement and plot the results obtained with an ``All Words" baseline, which performs geolocation classification by using all the words in the input text.

The IGR performance is very similar to the one reported in \cite{Han14} (45\% geolocation accuracy on Twitter for 100-mile-radius regions). The correlation of the performance gain with the number of features is almost identical, although as expected the absolute numbers are somehow higher on our dataset, since  the state granularity is somewhat easier to predict in comparison to the 100-mile-radius granularity used in \cite{Han14}.

We notice that in both datasets we can significantly improve the IGR performance by either using the WLH or the lexicon features. We believe this is an important result, given that IGR was previouly reported to lead to the best results for geolocation \cite{Han14}. Moreover, this improvement is achieved with significantly less features, which enables faster prediction models. We also notice that the improvement is more substantial in our Blogger dataset, which potentially suggests that WLH and lexicon features become increasingly more valuable as the number of words in a social media dataset increases.

Based on these evaluations on development data, we find that the best performing features for the geolocation of blogs are the top 60\% features ranked by our WLH heuristic
in conjunction with the LibLinear classifier. In contrast, on the Twitter dataset, the setting that works best is our WLH heuristic using 30\% of the features with a NB classifier.  

\mysubsection{Evaluations on the test data.} 
Based on these evaluations on development data, we find that the best performing features for the geolocation of blogs are the top 60\% features ranked by our WLH heuristic
in conjunction with the LibLinear classifier. In contrast, on the Twitter dataset, the setting that works best is our WLH heuristic using 30\% of the features with a NB classifier. Using these settings, we perform evaluations on the test dataset.

\begin{table}[htbp]
\centering
\scalebox{0.8}{
\begin{tabular}{  l | l | l }
\hline
Method & Blogger & Twitter \\
\hline 
Baseline (all words) & 21.68\% & 42.14\% \\ 
IGR (NB) \cite{Han14} & 23.1\% & 53.2\% \\ 
IGR (LibLinear) & 29.18\% & 49.65\% \\ 
WLH & \textbf{32.72\%*} &  \textbf{57.47\%*}\\ 
Lexicons & 31.1\% & 54.8\% \\ 
\hline 
Near-miss accuracy & 42.9\% & 66.51\% \\ 
\hline
\end{tabular}
}
\vspace{0.1in}
\caption{Geolocation accuracy on the Blogger and Twitter test data. Near-miss accuracy is reported for the best performing methods (WLH and LibLinear for blogs; WLH and NB for tweets).
\label{tab:acc}}
\end{table}

Table \ref{tab:acc} shows both accuracy, which measures the percentage of correct geolocation predictions on the test data, as well as the near-miss accuracy, which considers the prediction of a neighbouring state (states with common borders) also correct. 
The results obtained with our proposed feature selection methods are significantly better than those obtained with the IGR method.\footnote{Throughout the paper,  (*) and (**) denote a statistically significant difference using a 2-sample test with a p-value $<$ 0.01 and p-value $<$ 0.035 respectively.}

These results indicate that content-based geolocation is significantly more difficult for blogs. We believe this is an interesting finding, in particular since intuitively one would think that the length of the blogs (as compared to tweets) would help with this prediction task. One possible explanation as to why text from Twitter is easier to geolocate is the way users use this media: tweets, albeit shorter, appear to contain more location revealing words.

To further explore this indication, we apply a named entity recognizer (NER) on texts from both Blogger and Twitter. Since no social media specific NER tagger exists, we use the Stanford NER \cite{Finkel05} and tag the texts from 1,000 users from the training set of each platform.  Interestingly, we find that approximately 0.0076\% of the words are tagged as location names in Blogger while 0.0118\% are tagged as such in Twitter. The difference is statistically significant (p$<$2.2e-16) and could account to some extent for the difference in geolocation performance in the two media.

\mysubsection{Classifier efficiency.} Finally, in Table \ref{tab:time} we present the training and test time of the different geolocation classifiers. As seen in the table, the most efficient methods are the ones based on lexicons, followed by the WLH feature selection method with 30\% of the features. Even though the lexicons features give slightly lower performance from our best performing classifiers they can be incorporated to create models that are orders of magnitude faster that other approaches.

\begin{table*}[htbp]
\centering
\scalebox{0.8}{
\begin{tabular}{ l | l | c | c | c }
\hline
Features & Classifier & Media & Train Time (ms) & Test Time (ms) \\
\hline 
IGR (85\% top features) \cite{Han14} & NB & Blogger & 1,432 & 292 \\
IGR (65\% top features) \cite{Han14} & NB & Twitter & 429 & 118 \\
IGR (85\% top features) & LibLinear & Blogger & 605,514 & 110 \\
IGR (65\% top features) & LibLinear & Twitter & 153,948 & 64 \\
WLH (60\% top features) & LibLinear &  Blogger & 555,516 & 76 \\
WLH (30\% top features) & NB &  Twitter & 262 & 82 \\
Lexicons ($p=500,h=17,t=3$) & NB & Blogger & \textbf{141} & \textbf{53} \\
Lexicons ($p=11,h=16,t=2$) & NB & Twitter & \textbf{127} & \textbf{46} \\
\hline
\end{tabular}
}
\vspace{0.1in}
\caption{Geolocation training and test time as measured on our training and development sets.
\label{tab:time}}
\end{table*}

\section{Cross-media Geolocation}
\label{sec:crossgeo}

In addition to exploring content-based geolocation for individual media, our dual dataset of blogs and tweets also allows us to explore cross-media and multi-media geolocation. Since no single feature selection method was found to work best for both social media, in all the experiments reported in this section we once again identify the best settings by using a development set, and report the results obtained on a test set.

\subsection{Cross-media Predictions}
\label{subsec:crosspred}

To measure the role played by the social media type when training a geolocation model, we compare the results of the geolocation classifier when trained on blogs and tested on tweets, and vice versa when trained on tweets and tested on blogs. We further differentiate between the type of social media used to tune (develop) the system. For instance, when trained on blogs, the system can be tuned on blogs, and then applied on tweets; or it can be tuned on tweets, and then applied on tweets. 

\begin{table}[htbp]
\begin{center}
\scalebox{0.8}{
\begin{tabular}{ l l l l }
\hline
Training & Development & Test & Accuracy\\
\hline
Twitter  & Blogger & Blogger & 30.58\%\\
Twitter  & Twitter & Blogger & 27.28\%\\
{\it Blogger} & {\it Blogger} & {\it Blogger} & {\it 32.72\%}\\
\hline
Blogger  & Blogger & Twitter & 44.18\%\\
Blogger  & Twitter & Twitter & 44.02\%\\
{\it Twitter} & {\it Twitter} & {\it Twitter} & {\it 57.47\%} \\
\hline
\end{tabular}
}
\vspace{0.1in}
\caption{Cross-media geolocation. Within-media geolocation is also shown (in italic) to facilitate the comparisons.\label{tab:cross}}
\end{center}
\end{table}

Table \ref{tab:cross} shows the results obtained during these experiments. To facilitate the comparison with the within-media evaluations, the table also replicates the results reported in Table  \ref{tab:acc} (shown here in italic). Perhaps not surprisingly, the type of media that a system is trained on has a significant impact on the results. In the case of blogs, training on Twitter data results in a drop in accuracy of 3.3\%* absolute as compared to the case when the classifier is trained on Blogger data. An even bigger drop is noticed in the classification of tweets, where the change in the social media type used for training causes an accuracy loss of 17.5\%* absolute. Interestingly, the type of social media used for development has very little impact on performance (0.5\% absolute), and the size of the effect is consistent for both blogs and tweets.  

\subsection{Mixed-Media Prediction}
\label{subsec:mixed}

After exploring cross-media geolocation prediction, a natural follow-up question is whether we can improve the performance of a geolocation classifier by growing the training data with mixed media. Table \ref{table:aug} shows the geolocation results when training the classifier on a dataset consisting of the joint Blogger and Twitter training sets. In both evaluations, the development dataset belongs to the same social media as the test data.\footnote{Although, based on the results reported in Table \ref{tab:cross}, we would not expect significant differences if the development data were to be drawn from a different media.} As before, for comparison purposes, we also show (in italic) the results of the within-media evaluations from Table \ref{tab:acc}. 

\begin{table}[htbp]
\begin{center}
\scalebox{0.8}{
\begin{tabular}{ l l l l}
\hline
Training & Development & Test & Accuracy\\
\hline
Blogger+Twitter & Blogger & Blogger & 34.61\%**\\
{\it Blogger} & {\it Blogger} & {\it Blogger} & {\it 32.72\%}\\
\hline
Blogger+Twitter & Twitter & Twitter & 52.19\%\\
{\it Twitter} & {\it Twitter} & {\it Twitter} & {\it 57.47\%} \\
\hline
\end{tabular}
}
\vspace{0.1in}
\caption{Augmented training data from multiple sources.\label{table:aug}}
\end{center}
\end{table}

The geolocation of blogs appears to benefit from the augmentation of the training data with tweets, whereas the gelocation of tweets is worsened by the addition of blogs. This effect may be explained by our earlier observation that tweets are easier to geolocate, and therefore the addition of tweets to the training data leads to better features/lexicons, which is not the case when blogs are added to the training dataset of tweets.

\section{Geolocatability}
\label{sec:meta}
Motivated by the difference in geolocatability in the two social media, we explore this phenomenon further, and measure how certain demographics affect the geolocatability of text. All the results in this section are obtained by measuring the accuracy of the fully trained classifier (i.e., using the entire training set), tuned on the entire development dataset, and applied on a subset of the test set filtered for the selected demographic.

\subsection{State Geolocatability}

\begin{figure*}
\small
\begin{tabular*}{1\textwidth}{@{\extracolsep{\fill}} c c }
\multicolumn{1}{c}{\includegraphics[width=0.40\textwidth]{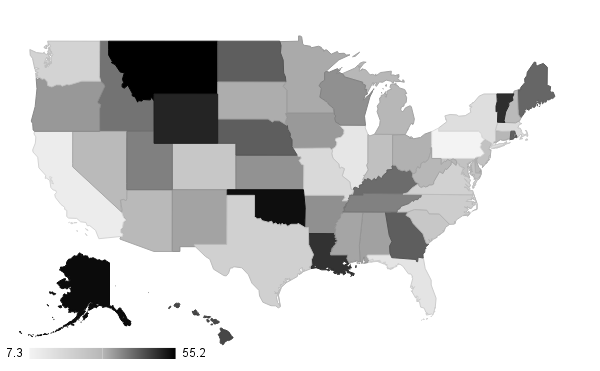}} &\multicolumn{1}{c}{\includegraphics[width=0.40\textwidth]{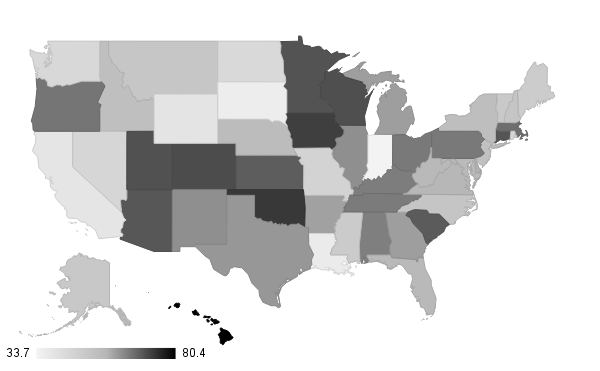}}\\
\multicolumn{1}{c}{Blogger} & \multicolumn{1}{c}{Twitter}\\
\end{tabular*}
\caption{Geolocatability across the 50 U.S. states for the two social media.} 
\label{fig:geoloc}
\end{figure*}

Figure \ref{fig:geoloc} shows the percentage of users in each state correctly geolocated, for both blogs and tweets. Interestingly, different states have significantly different geolocability, with users from  e.g., CA being harder to geolocate than users from e.g., OK. This could be attributed to the diversity of interests in highly-populated states such as CA, where the users speak less  about the location and more about other topics of interest, as well as with the popularity of many locations in these states (e.g., San Francisco) which are frequently mentioned by people outside the state, thus making the geolocability of these states harder.  

We also notice differences across the two media. While some states are consistently harder to geolocate in both media (e.g., CA, WA), others are easier to geolocate in Blogger (e.g. MN, AK), and others in Twitter (e.g., HI, IA). In fact, the Spearman correlation $\rho$ among the geolocatability distributions in the two media is 0.16 with a  p-value $<$ 0.25, which is not statistically significant \footnote{Using a different learning model (e.g., NB instead of LibLinear and vice versa) on the data of any media results in statistically significantly correlated distributions.
This suggests that the difference of the distributions between the two media noted above is not depended on the learning model used (i.e. LibLinear on Blogger and NB on Twitter)}. 
This suggests an even bigger gap in geolocability between blogs and tweets, adding to the overall difference noted in Section \ref{sec:detection}.

\subsection{Gender Geolocatability}
\label{subsec:metageo}

We also measure the geolocatability of the users based on their gender. We do this analysis only for the blog dataset, since we do not have this information available for the Twitter users. Using the user-declared gender in the users' profile, in Table \ref{tab:genderacc} we  measure the proportion of users in the test set that is correctly geotagged by our best performing, content-based classifier.

Unlike a previously published study that found that males are easier to geolocate on a large Twitter dataset \cite{Pavalanathan15}, we do not observe the same tendency in our Blogger dataset.

\begin{table}[htbp]
\begin{center}
\scalebox{0.8}{
\begin{tabular}{ l | l }
\hline
Gender & Accuracy\\
\hline
Male        & 31.17\%\\
Female      & 31.03\%\\
Undefined   & 38.71\%*\\
\hline
\end{tabular}
}
\vspace{0.1in}
\caption{Blogger geolocation per gender.\label{tab:genderacc}}
\end{center}
\end{table}


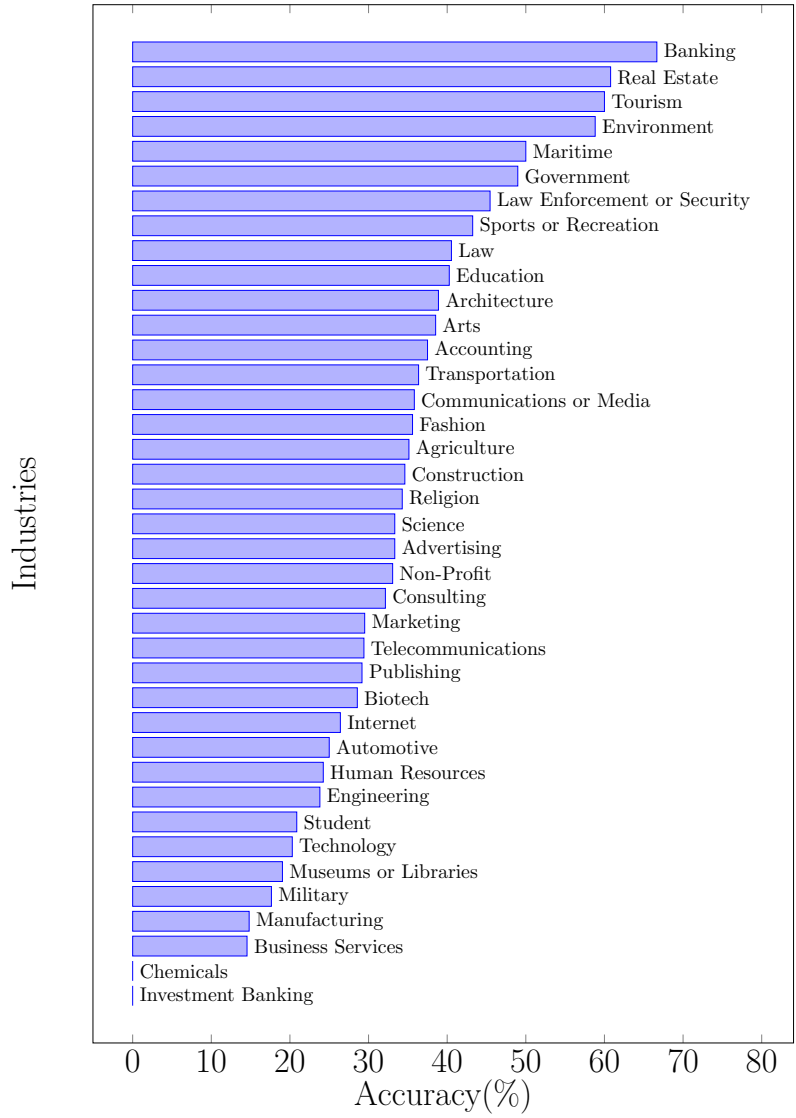
\begin{figure}
\vspace*{-1.5in}
\hspace*{-1.8in}
\scalebox{0.75}{

\begin{tikzpicture}
\begin{axis}[
xbar,
width=14cm,
height=20cm,
enlarge y limits={0.05},
xmin=-1,
xmax=80,
enlarge x limits={0.05},
xlabel={Accuracy(\%)},
ylabel={Industries},
ytick style={draw=none},
label style={font=\LARGE},
tick label style={font=\LARGE},
symbolic y coords={"Investment Banking","Chemicals","Business Services","Manufacturing","Military","Museums or Libraries","Technology","Student","Engineering","Human Resources","Automotive","Internet","Biotech","Publishing","Telecommunications","Marketing","Consulting","Non-Profit","Advertising","Science","Religion","Construction","Agriculture","Fashion","Communications or Media","Transportation","Accounting","Arts","Architecture","Education","Law","Sports or Recreation","Law Enforcement or Security","Government","Maritime","Environment","Tourism","Real Estate","Banking"},
ytick=data,
yticklabels={\textcolor{white}{"Investment Banking"},\textcolor{white}{"Chemicals"},\textcolor{white}{"Business Services"},\textcolor{white}{"Manufacturing"},\textcolor{white}{"Military"},\textcolor{white}{"Museums or Libraries"},\textcolor{white}{"Technology"},\textcolor{white}{"Student"},\textcolor{white}{"Engineering"},\textcolor{white}{"Human Resources"},\textcolor{white}{"Automotive"},\textcolor{white}{"Internet"},\textcolor{white}{"Biotech"},\textcolor{white}{"Publishing"},\textcolor{white}{"Telecommunications"},\textcolor{white}{"Marketing"},\textcolor{white}{"Consulting"},\textcolor{white}{"Non-Profit"},\textcolor{white}{"Advertising"},\textcolor{white}{"Science"},\textcolor{white}{"Religion"},\textcolor{white}{"Construction"},\textcolor{white}{"Agriculture"},\textcolor{white}{"Fashion"},\textcolor{white}{"Communications or Media"},\textcolor{white}{"Transportation"},\textcolor{white}{"Accounting"},\textcolor{white}{"Arts"},\textcolor{white}{"Architecture"},\textcolor{white}{"Education"},\textcolor{white}{"Law"},\textcolor{white}{"Sports or Recreation"},\textcolor{white}{"Law Enforcement or Security"},\textcolor{white}{"Government"},\textcolor{white}{"Maritime"},\textcolor{white}{"Environment"},\textcolor{white}{"Tourism"},\textcolor{white}{"Real Estate"},\textcolor{white}{"Banking"}},
nodes near coords,
nodes near coords align={horizontal},
nodes near coords={\pgfmathfloatifflags{\pgfplotspointmeta}{1}{}{}}
]
\addplot coordinates {(0,"Investment Banking") (0,"Chemicals") (14.5454545454545,"Business Services") (14.8148148148148,"Manufacturing") (17.6470588235294,"Military") (19.047619047619,"Museums or Libraries") (20.3007518796993,"Technology") (20.8695652173913,"Student") (23.8095238095238,"Engineering") (24.2424242424242,"Human Resources") (25,"Automotive") (26.4150943396226,"Internet") (28.5714285714286,"Biotech") (29.1666666666667,"Publishing") (29.4117647058824,"Telecommunications") (29.5081967213115,"Marketing") (32.1428571428571,"Consulting") (33.0645161290323,"Non-Profit") (33.3333333333333,"Advertising") (33.3333333333333,"Science") (34.28571429,"Religion") (34.61538462,"Construction") (35.13513514,"Agriculture") (35.59322034,"Fashion") (35.82089552,"Communications or Media") (36.36363636,"Transportation") (37.5,"Accounting") (38.53211009,"Arts") (38.88888889,"Architecture") (40.26622296,"Education") (40.54054054,"Law") (43.24324324,"Sports or Recreation") (45.45454545,"Law Enforcement or Security") (48.97959184,"Government") (50,"Maritime") (58.82352941,"Environment") (60,"Tourism") (60.7843137254902,"Real Estate") (66.6666666666667,"Banking")};
\addplot[black,sharp plot,update limits=false] coordinates {(0,"Investment Banking")} node[right] at (axis cs:0,"Investment Banking") {Investment Banking};
\addplot[black,sharp plot,update limits=false] coordinates {(0,"Chemicals")} node[right] at (axis cs:0,"Chemicals") {Chemicals};
\addplot[black,sharp plot,update limits=false] coordinates {(14.5454545454545,"Business Services")} node[right] at (axis cs:14.5454545454545,"Business Services") {Business Services};
\addplot[black,sharp plot,update limits=false] coordinates {(14.8148148148148,"Manufacturing")} node[right] at (axis cs:14.8148148148148,"Manufacturing") {Manufacturing};
\addplot[black,sharp plot,update limits=false] coordinates {(17.6470588235294,"Military")} node[right] at (axis cs:17.6470588235294,"Military") {Military};
\addplot[black,sharp plot,update limits=false] coordinates {(19.047619047619,"Museums or Libraries")} node[right] at (axis cs:19.047619047619,"Museums or Libraries") {Museums or Libraries};
\addplot[black,sharp plot,update limits=false] coordinates {(20.3007518796993,"Technology")} node[right] at (axis cs:20.3007518796993,"Technology") {Technology};
\addplot[black,sharp plot,update limits=false] coordinates {(20.8695652173913,"Student")} node[right] at (axis cs:20.8695652173913,"Student") {Student};
\addplot[black,sharp plot,update limits=false] coordinates {(23.8095238095238,"Engineering")} node[right] at (axis cs:23.8095238095238,"Engineering") {Engineering};
\addplot[black,sharp plot,update limits=false] coordinates {(24.2424242424242,"Human Resources")} node[right] at (axis cs:24.2424242424242,"Human Resources") {Human Resources};
\addplot[black,sharp plot,update limits=false] coordinates {(25,"Automotive")} node[right] at (axis cs:25,"Automotive") {Automotive};
\addplot[black,sharp plot,update limits=false] coordinates {(26.4150943396226,"Internet")} node[right] at (axis cs:26.4150943396226,"Internet") {Internet};
\addplot[black,sharp plot,update limits=false] coordinates {(28.5714285714286,"Biotech")} node[right] at (axis cs:28.5714285714286,"Biotech") {Biotech};
\addplot[black,sharp plot,update limits=false] coordinates {(29.1666666666667,"Publishing")} node[right] at (axis cs:29.1666666666667,"Publishing") {Publishing};
\addplot[black,sharp plot,update limits=false] coordinates {(29.4117647058824,"Telecommunications")} node[right] at (axis cs:29.4117647058824,"Telecommunications") {Telecommunications};
\addplot[black,sharp plot,update limits=false] coordinates {(29.5081967213115,"Marketing")} node[right] at (axis cs:29.5081967213115,"Marketing") {Marketing};
\addplot[black,sharp plot,update limits=false] coordinates {(32.1428571428571,"Consulting")} node[right] at (axis cs:32.1428571428571,"Consulting") {Consulting};
\addplot[black,sharp plot,update limits=false] coordinates {(33.0645161290323,"Non-Profit")} node[right] at (axis cs:33.0645161290323,"Non-Profit") {Non-Profit};
\addplot[black,sharp plot,update limits=false] coordinates {(33.3333333333333,"Advertising")} node[right] at (axis cs:33.3333333333333,"Advertising") {Advertising};
\addplot[black,sharp plot,update limits=false] coordinates {(33.3333333333333,"Science")} node[right] at (axis cs:33.3333333333333,"Science") {Science};
\addplot[black,sharp plot,update limits=false] coordinates {(34.28571429,"Religion")} node[right] at (axis cs:34.28571429,"Religion") {Religion};
\addplot[black,sharp plot,update limits=false] coordinates {(34.61538462,"Construction")} node[right] at (axis cs:34.61538462,"Construction") {Construction};
\addplot[black,sharp plot,update limits=false] coordinates {(35.13513514,"Agriculture")} node[right] at (axis cs:35.13513514,"Agriculture") {Agriculture};
\addplot[black,sharp plot,update limits=false] coordinates {(35.59322034,"Fashion")} node[right] at (axis cs:35.59322034,"Fashion") {Fashion};
\addplot[black,sharp plot,update limits=false] coordinates {(35.82089552,"Communications or Media")} node[right] at (axis cs:35.82089552,"Communications or Media") {Communications or Media};
\addplot[black,sharp plot,update limits=false] coordinates {(36.36363636,"Transportation")} node[right] at (axis cs:36.36363636,"Transportation") {Transportation};
\addplot[black,sharp plot,update limits=false] coordinates {(37.5,"Accounting")} node[right] at (axis cs:37.5,"Accounting") {Accounting};
\addplot[black,sharp plot,update limits=false] coordinates {(38.53211009,"Arts")} node[right] at (axis cs:38.53211009,"Arts") {Arts};
\addplot[black,sharp plot,update limits=false] coordinates {(38.88888889,"Architecture")} node[right] at (axis cs:38.88888889,"Architecture") {Architecture};
\addplot[black,sharp plot,update limits=false] coordinates {(40.26622296,"Education")} node[right] at (axis cs:40.26622296,"Education") {Education};
\addplot[black,sharp plot,update limits=false] coordinates {(40.54054054,"Law")} node[right] at (axis cs:40.54054054,"Law") {Law};
\addplot[black,sharp plot,update limits=false] coordinates {(43.24324324,"Sports or Recreation")} node[right] at (axis cs:43.24324324,"Sports or Recreation") {Sports or Recreation};
\addplot[black,sharp plot,update limits=false] coordinates {(45.45454545,"Law Enforcement or Security")} node[right] at (axis cs:45.45454545,"Law Enforcement or Security") {Law Enforcement or Security};
\addplot[black,sharp plot,update limits=false] coordinates {(48.97959184,"Government")} node[right] at (axis cs:48.97959184,"Government") {Government};
\addplot[black,sharp plot,update limits=false] coordinates {(50,"Maritime")} node[right] at (axis cs:50,"Maritime") {Maritime};
\addplot[black,sharp plot,update limits=false] coordinates {(58.82352941,"Environment")} node[right] at (axis cs:58.82352941,"Environment") {Environment};
\addplot[black,sharp plot,update limits=false] coordinates {(60,"Tourism")} node[right] at (axis cs:60,"Tourism") {Tourism};
\addplot[black,sharp plot,update limits=false] coordinates {(66.6666666666667,"Banking")} node[right] at (axis cs:66.6666666666667,"Banking") {Banking};
\addplot[black,sharp plot,update limits=false] coordinates {(60.7843137254902,"Real Estate")} node[right] at (axis cs:60.7843137254902,"Real Estate") {Real Estate};

\end{axis}
\end{tikzpicture}

}
\vspace*{-1.4in}
\caption{Industry geolocatability in blogs.\label{fig:geolocInd}}
\end{figure}

Surprisingly, the users who have not defined their gender are a lot easier to geolocate than males or females.

\subsection{Industry Geolocatability}

Another well defined element available in Blogger profiles is their industry. Once again, we perform this analysis on blogs only as we lack this information for the Twitter users. Figure \ref{fig:geolocInd} shows the geolocatability of Blogger users for the 39 different industries. With the exception of the \textit{Chemicals}  and \textit{Investment Banking} industries, which were underrepresented in our dataset (2-3 users), and hence prone to extreme accuracy results (e.g. 0\%), 
we find a large variation in the geolocatability of users based on industry.

For instance, users who are in the {\it Real Estate} and {\it Tourism} industries are the easiest to geolocate, perhaps because their work-related posts are more likely to include toponyms. On the other end, users that work in the {\it Architecture} and {\it Manufacturing} industries are particularly hard to geotag.
This result suggests an additional factor of the underlying population that uses a particular platform, which influences their geolocatability.

\section{Conclusion}
\label{sec:conclusion}

In this paper, we examined large-scale content-based geolocation on social media. Using two large comparable datasets of blogs and tweets, and two new feature selection approaches, we ran several experiments that allowed us to compare the performance of geolocation prediction using different media. 

The new lexicon features that we proposed brought a relative error rate reduction of up to 10.4\% for the geolocation of Blogger and Twitter users, as compared to a manual feature selection method found to work best in previous work \cite{Han14}. Similarly, the word locality heuristic (WLH) that we introduced brought a relative error rate reduction of 9.1\% in geolocation accuracy when compared  to the same previous method. 

Our findings also indicate that despite their longer text, Blogger users are significantly harder to geolocate. This result suggests that despite the focus of the current geolocation research on Twitter data,   the application of geospatial technologies on social media platforms other than Twitter will be more challenging.

We also experimented with cross-media classification, and showed that the media used for training does have an effect on the accuracy of the geolocation classifiers, with lower accuracy figures obtained when the training data is drawn from a  social media stream different from the test data. We also explored the use of mixed-media as a way to augment the training data, and found  that the geolocation of Blogger users can benefit from incorporating additional Twitter training data, but the same does not apply to the geolocation of Twitter users. To our knowledge, this is the first study that compares methods on different media.

Finally, an analysis of geolocatability based on user demographics showed that the state, industry, or gender of the users play a role in how easy (or difficult) it is to geolocate them. This points to a potential future research direction, with geolocation classifiers targeted to certain user dimensions.

To encourage more research on text-based geolocation on blog data, the code used to collect the Blogger data used in this study is publicly available at http://lit.eecs.umich.edu.

\section*{Acknowledgment}
This material is based in part upon work supported by the National Geospatial Agency (grant \#HM02101310006), by the National Science Foundation (grant \#1344257), and the John Templeton Foundation (grant \#48503). Any opinions, findings, and conclusions or recommendations expressed in this material are those of the authors and do not necessarily reflect the views of the National Geospatial Agency, the National Science Foundation, or the John Templeton Foundation.

\section*{}
\bibliographystyle{elsarticle-num}
\bibliography{cls}

\parpic{\includegraphics[width=1in,clip,keepaspectratio]{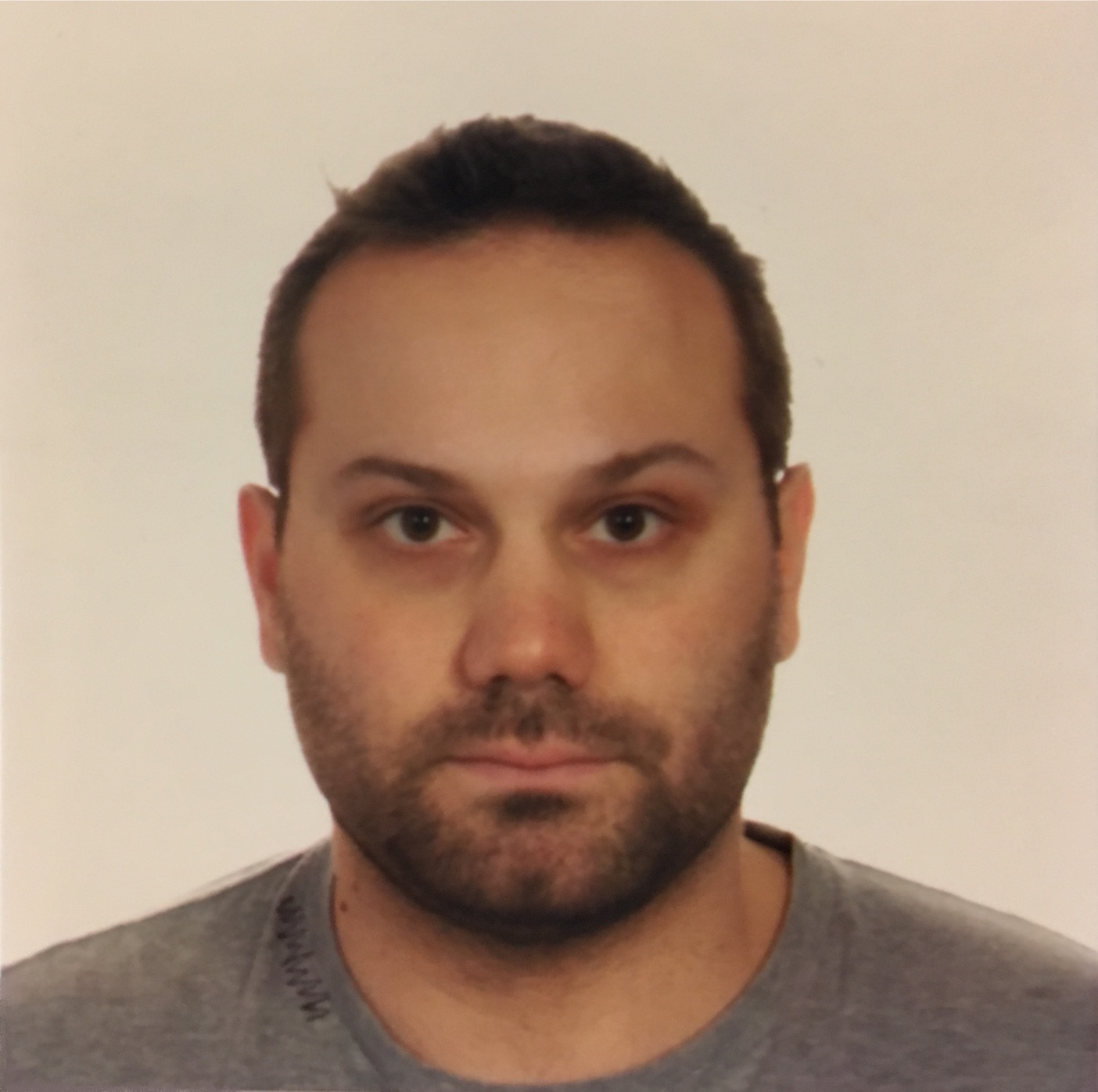}}
\noindent {\bf Konstantinos Pappas} received his Master of Science degree from the Department of Computer Science and Engineering at the University of Michigan (2016) and earned his Bachelor's degree at the Department of Informatics at the Athens University of Economics and Business (2009). His current research interests include intelligent systems, natural language processing, and applied machine learning. His contributions to the work presented in this paper were made while he was a PhD Candidate in the Department of Computer Science and Engineering at the University of Michigan before his affiliation with Amazon.com.

\parpic{\includegraphics[width=1in,clip,keepaspectratio]{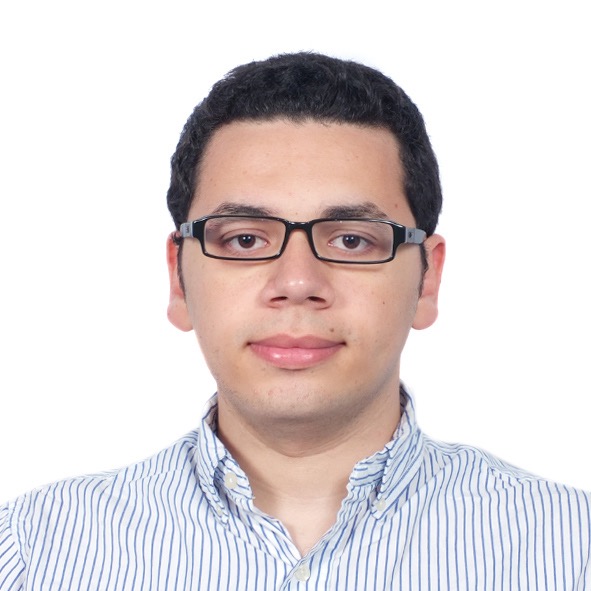}}
\noindent {\bf Mahmoud Azab} is a PhD Candidate in the Department of Computer Science and Engineering at the University of Michigan. He received his bachelor's degree from Cairo University in 2011. Prior to joining the University of Michigan, he worked as a research assistant for two years at Carnegie Mellon University in Qatar. His research areas of interest include natural language processing and multimodal machine learning.

\parpic{\includegraphics[width=1in,clip,keepaspectratio]{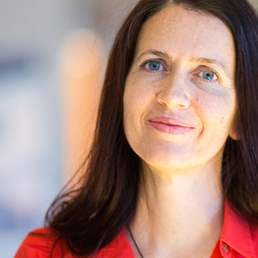}}
\noindent {\bf Rada Mihalcea} is a  Professor in the Department of Computer Science and Engineering at the University of Michigan. Her research interests are in computational linguistics, with a focus on lexical semantics, graph-based algorithms for natural language processing, and multilingual natural language processing. She serves or has served on the editorial boards of the Journals of Computational Linguistics, Language Resources and Evaluations, Natural Language Engineering, Research in Language in Computation, IEEE Transactions on Affective Computing, and Transactions of the Association for Computational Linguistics. She was a program co-chair for the Conference of the Association for Computational Linguistics (2011) and the Conference on Empirical Methods in Natural Language Processing (2009), and a general chair for the Conference of the North American Association for Computational Linguistics (NAACL 2015). She is the recipient of a National Science Foundation CAREER award (2008) and a Presidential Early Career Award for Scientists and Engineers (2009).


\end{document}